%% file: main.tex
\documentclass[10pt,twocolumn,letterpaper]{article}

\usepackage[pagenumbers]{cvpr} % To force page numbers, e.g. for an arXiv version


\definecolor{cvprblue}{rgb}{0.21,0.49,0.74}
\usepackage[pagebackref,breaklinks,colorlinks,allcolors=cvprblue]{hyperref}

\usepackage{multirow}
\definecolor{firstBest}{rgb}{0.86, 1, 0.86} % green
\usepackage{xcolor}
\usepackage{colortbl}
\usepackage{pifont} % For checkmark and cross mark
\newcommand{\cmark}{\ding{51}} 
\newcommand{\xmark}{\ding{55}} 
\usepackage{booktabs}
\usepackage{bm}

\def\paperID{*****} % *** Enter the Paper ID here
\def\confName{CVPR}
\def\confYear{2025}

\title{CombatVLA: An Efficient Vision-Language-Action Model for Combat Tasks in 3D Action Role-Playing Games}

\author{\textbf{Peng Chen*}, \textbf{Pi Bu*}, \textbf{Yingyao Wang}, \textbf{Xinyi Wang}, \textbf{Ziming Wang}, \textbf{Jie Guo}, \\ 
\textbf{Yingxiu Zhao}, \textbf{Qi Zhu}, \textbf{Jun Song$\dagger$}, \textbf{Siran Yang}, \textbf{Jiamang Wang}, \textbf{Bo Zheng} \\
Alibaba Group \\
\{zhaojun.cp, bupi.wj, jsong.sj\}@taobao.com
}

\begin{document}
\maketitle

\renewcommand{\thefootnote}{\fnsymbol{footnote}} 
\footnotetext[1]{Equal Contribution.}
\footnotetext[2]{Corresponding Author.}

\input{sec/0_abstract}    
\input{sec/1_intro}
\input{sec/2_related}
\input{sec/3_method}

\input{sec/4_experiment}

\input{sec/5_conclusion}

% \clearpage
% \newpage
{
    \small
    \bibliographystyle{ieeenat_fullname}
    \bibliography{main}
}

% \appendix
\input{sec/X_supple}

\end{document}

%% file: sec/0_abstract.tex
\begin{abstract}
Recent advances in Vision-Language-Action models (VLAs) have expanded the capabilities of embodied intelligence. However, significant challenges remain in real-time decision-making in complex 3D environments, which demand second-level responses, high-resolution perception, and tactical reasoning under dynamic conditions. To advance the field, we introduce CombatVLA, an efficient VLA model optimized for combat tasks in 3D action role-playing games(ARPGs). Specifically, our CombatVLA is a 3B model trained on video-action pairs collected by an action tracker, where the data is formatted as action-of-thought (AoT) sequences. Thereafter, CombatVLA seamlessly integrates into an action execution framework, allowing efficient inference through our truncated AoT strategy. Experimental results demonstrate that CombatVLA not only outperforms all existing models on the combat understanding benchmark but also achieves a 50-fold acceleration in game combat. Moreover, it has a higher task success rate than human players. We will open-source all resources, including the action tracker, dataset, benchmark, model weights, training code, and the implementation of the framework at \url{https://combatvla.github.io/}.

\end{abstract}

%% file: sec/1_intro.tex
\section{Introduction}
% \label{sec:intro}
\begin{figure}[!ht]
    \centering
    \includegraphics[width=1.0\linewidth]{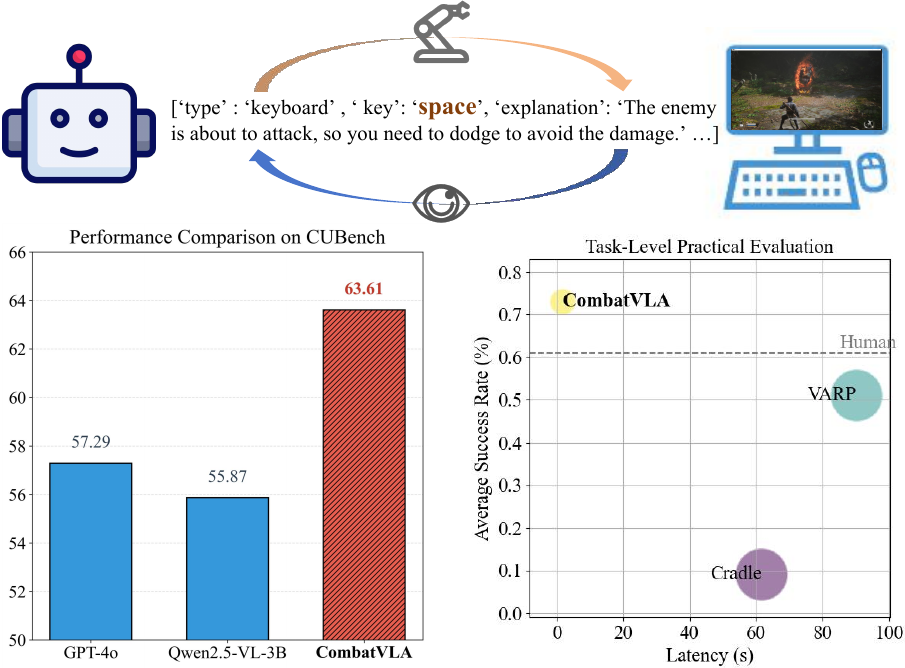}
    \vspace{-0.3cm}
    \caption{\textbf{CombatVLA surpasses GPT-4o and Qwen2.5-VL in combat understanding, is 50 times faster than Cradle and VARP framework, and has a higher success rate than humans.}
    %Our CombatVLA can efficiently perform combat tasks in 3D ARPGs, achieving state-of-the-art performance on CUBench and task-level practical tests.
    }
    \label{fig:intro}
    \vspace{-0.5cm}
\end{figure}

% Background
Vision-Language-Action Models (VLAs) have achieved groundbreaking progress in embodied intelligence through unified frameworks integrating visual perception, semantic reasoning, and physical action control \cite{brohan2023rt2,kim2024openvla}. In agent applications, they perform well in UI operations and navigation tasks \cite{lin2024showui,li2025hedgeagents} but \textbf{struggle with efficient decision-making in complex 3D environments}. A representative challenge lies in combat tasks in 3D action role-playing games(ARPGs) like ``Black Myth: Wukong'', which present critical but underexplored demands: real-time processing of high-resolution visual streams, tactical adaptation to dynamically evolving enemy behaviors, and second-level action execution—requirements mirroring latency-sensitive real-world scenarios \cite{turing2024aiwukong,chen2024varp}.  
% the importance of 3D combat games for research
In fact, the dynamic complexity of combat games
%—visually diverse enemies, multifaceted attack patterns, intricate 3D environments with dynamic effects, and sub-second evasion/attack windows—
rigorously challenges VLAs' capabilities in:  
1) Visual perception (e.g., enemy and self positioning, movement, and environmental awareness).
2) Combat reasoning (e.g., recognizing enemy attack patterns).
% 3) Precision action control (e.g., coordinating multiple action sequences).
3) Efficient inference (i.e., real-time reaction).
Currently, no framework excels in these tasks, nor is there a benchmark for assessing combat understanding.
%We believe that 3D combat games serve not only as stress-testing platforms for VLA capabilities but also as frontier simulation environments for embodied AI and "ultimate proving grounds" for complex decision-making tasks. 
%We believe 3D combat games are ideal for testing VLA capabilities and simulating real-world complex decision-making tasks.

\begin{figure*}[h]
\vspace{-12pt}
\centering
\includegraphics[width=1.0\textwidth]{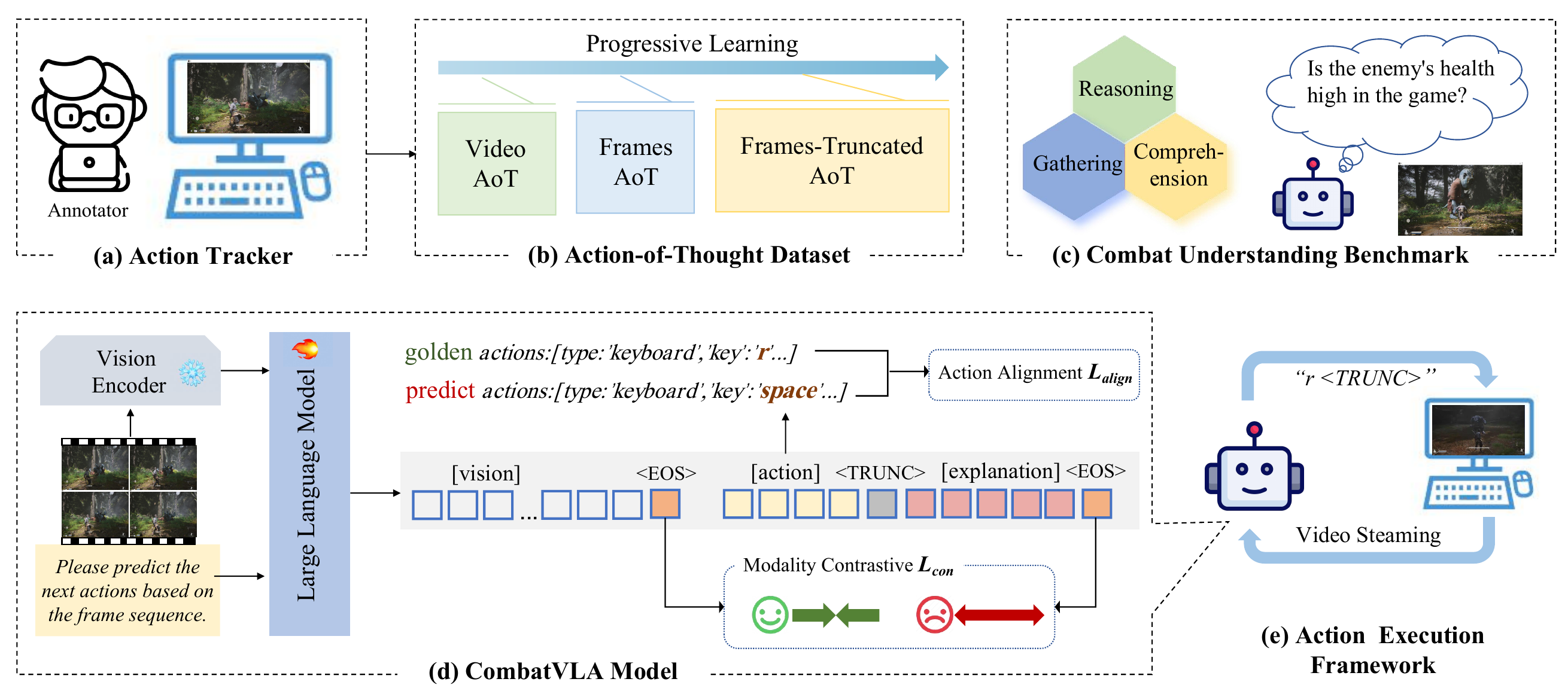}
\vspace{-0.8cm}
\caption{\textbf{(a) An action tracker} collects human data on keyboard and mouse use. \textbf{(b) Three types of AoT training data} collected by the action tracker are used for progressive learning. \textbf{(c) Combat understanding benchmark (namely CUBench)} assesses the model's combat IQ in three areas: gathering, comprehension, and reasoning. \textbf{(d) CombatVLA model} is trained on AoT data with the constraint of action alignment loss and modality contrastive loss. \textbf{(e) Deployment of CombatVLA} to operate real PCs.}
\label{fig:framework}
\vspace{-0.4cm}
\end{figure*}

% limitations
The pioneers working on 3D combat games primarily access video game APIs to read memory and thereby acquire information on the game environment. For example, \citet{wang2023voyager} used this strategy to employ LLM-driven agents to play Minecraft, enabling automatic mining, exploration, and combat with enemies. However, this method of interacting with the environment is significantly different from that of humans, who rely on vision rather than memory-based reading. %Furthermore, most games do not support API access. 
Recently, \cite{turing2024aiwukong} implemented reinforcement learning (RL) to play ``Black Myth: Wukong,'' using DQN and PPO algorithms with pure visual input, allowing AI to autonomously learn combat scenarios. However, this RL-based approach requires a large number of pre-defined reward designs and extensive trial-and-error training. %The trained AI exhibits poor generalization capabilities and can easily fail when faced with new opponents. 
With the advancement of visual language models (VLMs) \cite{yang2023auto, shinn2023reflexion}, works like Cradle \cite{tan2024cradle} and VARP \cite{chen2024varp} demonstrate significant potential in playing video games. Nonetheless, these efforts heavily depend on ultra-large-scale VLMs like GPT-4o, leading to delays that can exceed 60 or even 90 seconds, as shown in Fig.\ref{fig:intro}. This latency severely hinders the performance in real-time combat games and limits the practical applicability.

In this paper, we propose \textbf{CombatVLA}, the first efficient visual-language action model designed for 3D combat gameplay. For efficient decision making, our CombatVLA is a 3B model that processes visual inputs and outputs a sequence of actions to control the game (including keyboard and mouse operations). Specifically, we first develop an action tracker to collect a substantial amount of training data. The data gathered by this tracker is then structured into an action-of-thought (AoT) format to facilitate action reasoning by the model. Thereafter, CombatVLA is trained using a progressive learning paradigm, enabling the model to learn combat techniques, from video-level AoT tuning to frame-level AoT tuning. Ultimately, CombatVLA can be seamlessly integrated into an action execution agent, enabling efficient inference through our custom truncated output strategy. As shown in Fig.\ref{fig:intro}, the experimental results demonstrate that CombatVLA not only outperformed all existing models (e.g., GPT-4o and Qwen2.5-VL) in the combat understanding but also achieved a 50-fold increase in execution speed compared to existing VLM-based game agents.
The contributions are summarized as follows:
\begin{itemize}[leftmargin=*]
 \item{\textbf{Action Tracker.} We develop an action tracker that operates in the background of the game to record the player’s movements. This tool will expedite data collection, potentially advancing research in the field.}
 \item{\textbf{Benchmark of Combat Understanding.} Based on the action tracker, we establish a benchmark, namely CUBench, for combat understanding that evaluates the models' performance in identifying enemy positions and action reasoning tasks through a VQA format.}
 \item{\textbf{AoT Dataset.} We introduce a three-stage AoT dataset consisting of coarse-grained video AoT, fine-grained frames AoT and frames-truncated AoT, to enable the model to progressively learn combat skills.}
 \item{\textbf{CombatVLA Model.} CombatVLA is trained using a progressive learning paradigm, with constraints imposed by adaptive action-weighted losses, and it achieves optimal performance on combat understanding benchmark.}
 \item{\textbf{Action Execution Framework.} We integrate CombatVLA into an agent framework that operates on PCs, achieving a 50-fold acceleration via truncated strategies.}
 % \item{\textbf{All Resources.} We will open-sourcing all resources, including the dataset, action tracker, model weights, training code, and framework implementation.}
% \item{\textbf{Experiments.} Comprehensive experiments demonstrate that our approach achieves optimal performance in tasks within the 3D combat gameplay scenario.}
\end{itemize}

%% file: sec/2_related.tex
\section{Related Work}

\begin{figure*}[h]
\vspace{-0.2cm}
\centering
\includegraphics[width=1.0\textwidth]{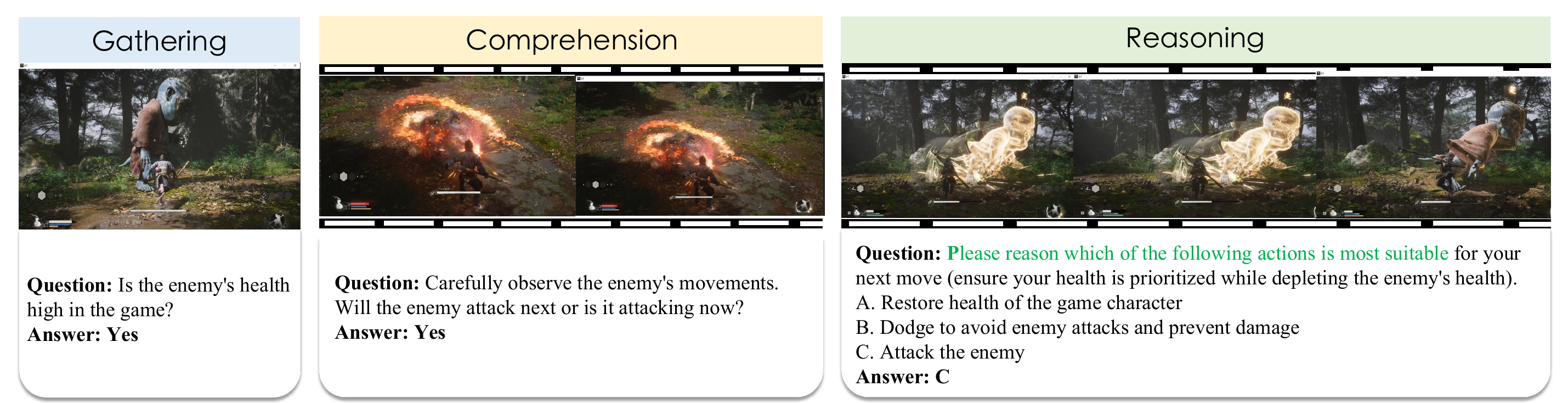}
\vspace{-0.8cm}
\caption{Combat understanding benchmark (i.e. CUBench) has three categories: gathering (single-image judgment), comprehension (multi-image judgment), and reasoning (multiple-image multiple-choice).}
\label{fig:benchmark}
\vspace{-0.4cm}
\end{figure*}

\subsection{Vision-Language-Action Models}
%The field of embodied AI has witnessed significant advances in vision-language-action (VLA) models, which unify three components: visual encoders~\citep{resnet, cnn, vit}, language encoders~\citep{llm}, and action decoders~\citep{rl}. Current VLA research focuses on three areas: multimodal pretraining[], control policy optimization[], and task planning[].
% 
% VLAs utilizing LLM-based control policies demonstrate strong generalization. RT-2~\citep{rt2} combines ViT-based vision with PaLM's language reasoning for robotic control. OpenVLA~\citep{openvla} enhances generalization via large-scale visuomotor pretraining. DeeR-VLA~\citep{deervla} reduces LLM computation with dynamic capacity adjustment, while RoboFlamingo~\citep{roboflamingo} separates VLM policies into vision-language and action modules for resource-limited platforms.
% % 
% In graphical interface automation, ShowUI~\citep{showui} and PPTAgent~\citep{pptagent} process high-resolution screenshots using UI-guided token selection and vision-language-action streaming pipelines. 
% % 
% However, current VLA methods still face challenges in achieving real-time responsiveness for latency-sensitive applications, particularly when executing long-horizon planning in complex 3D environments with dynamic visual effects and sub-second action windows.

With the development of Vision-Language Models (VLMs), several robust models, such as the Qwen series~\cite{wang2024qwen2}, have demonstrated strong visual capabilities~\cite{gu2025chinesesimplevqa}. Subsequently, VLMs were extended to Vision-Language Agents (VLAs) to further advance the development of embodied intelligence~\cite{guo2025vla_rl}. VLAs that utilize LLM-based control strategies exhibit strong generalization capabilities~\cite{cheang2024gr2,pan2023llms}. For instance, RT-2 \citep{brohan2023rt2} integrates VIT-based vision with PaLM's linguistic reasoning for robot control, while OpenVLA \citep{kim2024openvla} enhances generalization through large-scale visual-motor pre-training. DeeR-VLA \citep{yue2024deervla} reduces LLM computation through dynamic capacity adjustment, and RoboFlamingo \citep{li2024robotflamingo} separates VLM strategies into distinct vision-language and action modules, making it suitable for resource-limited platforms.
However, current VLA approaches still face challenges in achieving real-time response for latency-sensitive applications, particularly when executing long-horizon planning in complex 3D environments with dynamic visual effects and second-level action windows.

\subsection{AI-Driven Game Agents}
% The evolution of game agents follows two complementary paradigms: traditional reinforcement learning (RL)-based approaches and modern LLM-driven cognitive architectures.
% %
% RL-based methods achieve proficiency in specific scenarios through meticulous reward engineering. The Black Myth: Wukong AI initiative~\citep{blackmythai} implements pure vision-based DQN/PPO algorithms for action RPGs. While JARVIS-1~\citep{jarvis} and VPT~\citep{vpt} attempt human-like interaction through raw screenshot inputs and keyboard/mouse controls, their game-specific action space definitions limit generalization. Cradle~\citep{tan2024cradle} implements universal computer control, enabling agents to process multimodal inputs, execute decision-making, and interact with computer tasks without dedicated APIs. A memory module stores and maintains all relevant information acquired from both environmental interactions and LLM outputs. However, these methods require extensive environmental feedback collection and exhibit limited adaptability to novel opponents.
% %
% LLM-driven agents leverage language models' reasoning capabilities in turn-based board games and text adventures \citep{calm,cicero}. Minecraft agents like voyager \cite{wang2023voyager} demonstrate GPT-4's code generation capabilities through integration with the Mineflayer API. Recent LLM-based agents such as PokeLLMon~\citep{pokellmon} exemplify the potential of LLMs as decision-makers by improving understanding and strategic capabilities in Pokémon battles.
%

The development of game agents involves two main approaches: RL- and LLM-based architectures~\cite{deng2023mind2webgeneralistagentweb, gur2024realworldwebagentplanninglong, he2024webvoyagerbuildingendtoendweb, wang2024mobileagentautonomousmultimodalmobile, zhang2023appagentmultimodalagentssmartphone,qian2024chatdev, park2023generative,cai2025rtbagentllmbasedagentrealtime,zhai2024finetuninglargevisionlanguagemodels,ellis2024smacv2}. RL-based methods excel in specific tasks through reward engineering. The project of Black Myth: Wukong AI~\citep{turing2024aiwukong} uses vision-based DQN/PPO algorithms for action RPGs. JARVIS-1~\citep{wang2023jarvis1} and VPT~\citep{jucys2024vpt} mimic human interactions using screenshots and keyboard/mouse inputs, but struggle with generalization due to predefined action spaces. 
LLM-driven agents use language models for reasoning in board games and text adventures. Minecraft agents like Voyager~\cite{wang2023voyager} show GPT-4's ability to generate code. Agents such as PokeLLMon~\citep{hu2024pokellmon} enhance understanding and strategy in Pokémon battles, demonstrating LLMs' decision-making potential. Cradle~\cite{tan2024cradle} offers universal computer control without dedicated APIs, but needs extensive feedback and is less adaptable to new tasks. 

%Our work addresses this gap by integrating LLMs' reasoning with VLA models' visuomotor precision, creating a novel agent class that maintains human-like responsiveness while adapting to dynamic 3D environments, enabling real-time decision-making in vision-only game settings. The 3D combat environment serves as an ultimate proving ground, rigorously testing VLA capabilities across three dimensions: high-dimensional visual perception , semantic combat logic understanding, and second-level action control.

%% file: sec/3_method.tex
\section{Tracker and Benchmark}
%To advance the AI-based combat game models, 
We develop an action tracker to collect human action sequences in games. It provides extensive training data for a combat understanding model. Moreover, we establish a comprehensive combat understanding benchmark, with three tasks using the action tracker.

\begin{figure*}[h]
\vspace{-0.2cm}
\centering
\includegraphics[width=1.0\textwidth]{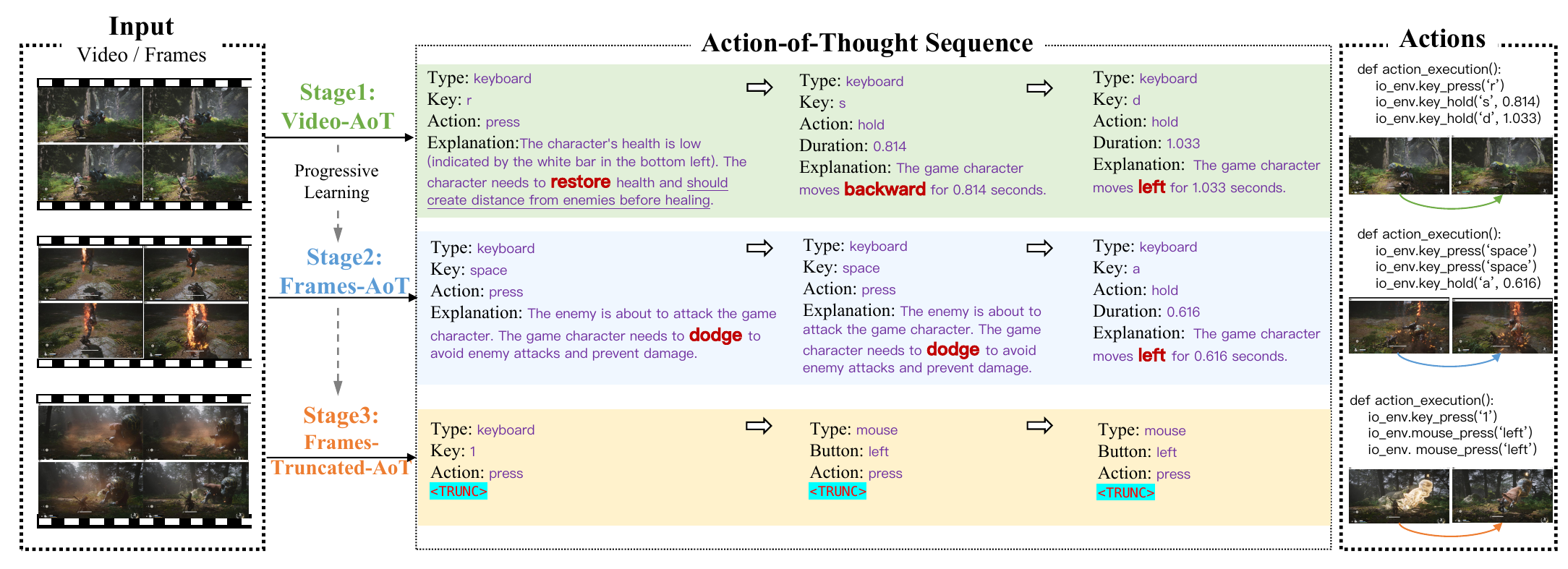}
\vspace{-0.8cm}
\caption{\textbf{Visualization of action-of-thought reasoning.} Given a video or frame input, the model can quickly infer the correct action under the semantic guidance of AoT. The special token $\langle \text{TRUNC} \rangle $ speeds up efficient reasoning by truncating output.}
\label{fig:AoT}
\vspace{-0.4cm}
\end{figure*}

\paragraph{Action Tracker.}
Due to the scarcity of training data labeled with actions, we have developed a lightweight python tool for efficiently collecting video-action data pairs, called the action tracker. It can run in the background, monitoring keyboard and mouse activities to record the user's actions, while simultaneously capturing game screenshots. Since it uses two separate threads, it is necessary to record timestamps to align frames with actions.
The frame set \( F = \{ f_1, f_2, \ldots, f_N \} \) consists of frames, where each frame \( f_i \) is associated with a timestamp \( t_{f_i} \), and they are ordered such that \( t_{f_1} \leq t_{f_2} \leq \cdots \leq t_{f_N} \). Similarly, the action set \( A = \{ a_1, a_2, \ldots, a_M \} \) comprises actions, where each action \( a_j \) is linked with a timestamp \( t_{a_j} \), and they are arranged in a sequence such that \( t_{a_1} \leq t_{a_2} \leq \cdots \leq t_{a_M} \). 
Then the alignment formula is as follows:
\begin{equation}
\forall a_j \in A,  a_j \mapsto f_{i_j} 
\end{equation}
where $i_j = \underset{i}{\arg\min} \, (t_{f_i} \geq t_{a_j})$. It ensures that each action is aligned with the nearest future frame.

\paragraph{Combat Understanding Benchmark.}
If a VLM-based or VLA-based model is expected to perform well in 3D ARPGs, it needs to have high-dimensional visual perception, and semantic combat logic understanding. Therefore, we establish a combat understanding benchmark (namely, CUBench) centered around three capabilities (i.e. gathering, comprehension, and reasoning), as shown in Fig.\ref{fig:benchmark}, to assess the model's combat IQ.

The evaluation data are sourced from the action tracker, which collected recordings and screenshots of human actions in ``Black Myth: Wukong,'' a 3D action game. The human annotation team comprised six individuals, each of whom had completed all levels of the game. Over the course of two weeks, they labeled the game data, resulting in a collection of 200 hours of recorded actions\footnote{The cleaned video data will also be used for the AoT training set without overlap to ensure fair evaluation.}.

We then used GPT-4o-0513 to create QA pairs for three tasks: single-image judgment, multi-image judgment, and multiple-image multiple-choice. Notably, all QA pairs were annotated by a team of ten people and cross-verified to ensure the quality. Ultimately, we compiled 914 data pieces (39.4\% gathering, 22.3\% comprehension, 38.3\% reasoning) to test the model's combat understanding. All prompts and data analysis are in the supplementary material.
%, annotation requirements 

\section{CombatVLA}
As illustrated in Fig.\ref{fig:framework}, our CombatVLA is a 3B model designed for efficient inference, capable of processing visual inputs and producing a sequence of actions to control the game (including both keyboard and mouse operations). The training process of CombatVLA follows a three-step  progressive learning paradigm, progressing from video-level training to frame-level training. Ultimately, CombatVLA can be seamlessly integrated into the action execution framework, allowing for efficient inference through our truncated AoT strategy.
%CombatVLA, as shown in Fig.{xx}, is built upon the vision-language model Qwen2.5-VL-3B as the backbone, and incorporates the following key components: (1) Inspired by chain-of-thought (CoT), we constructed our dataset, collected using an action tracker, into an action-of-thought (AoT) format to enhance the model's understanding of actions; (2) Curriculum-learning action tuning allows the model to progressively learn combat paradigms, enabling it to temporally locate actions at the frame level; (3) We developed an action execution framework that allows the model to play games like a human and make rapid responses.

\subsection{Action-of-Thought Construction}
Chain-of-Thought (CoT) prompting has been proven extremely effective in enhancing the complex reasoning capabilities of LLMs and MLLMs \cite{wei2023cot,li2025vot}. Inspired by CoT, we transform the data collected from the action tracker—which includes the set of frames \( F \), the set of actions \( A \), and their alignment—into action-of-thought data (AoT), as illustrated in Fig.~\ref{fig:AoT}. Specifically, the model response is formatted in JSON, which includes [action] (such as ``press space'') and [explanation] (used to describe the current state of the enemy, the physical meaning of the action, etc.). Additionally, the special token $ \langle \text{TRUNC} \rangle $ represents output truncation to increase inference speed.
%Specifically, we structure actions into a JSON format and add an `explanation' key-value pair for each action. This explains the current state of the enemy or game character, the physical significance of the action, and the high-level semantic information about why this action needs to be performed. AoT is similar to human cognitive processes and can establish high-dimensional conditional connections between images and actions. For more details on AoT, please refer to the supplementary materials.
\vspace{-0.6cm}
\begin{center}
\fcolorbox{black}{gray!10}{\parbox{1\linewidth}{
\textbf{Question:} $\langle \text{IMG} \rangle $ $\langle \text{IMG} \rangle $ $\langle \text{IMG} \rangle $ Please predict the next actions based on the frame sequence. \\
\textbf{Answer:} [action] \ $ \langle \text{TRUNC} \rangle $ \ [explanation] \ $\langle \text{EOS} \rangle $.
}}
\end{center}

% \paragraph{Truncated AoT Construction.}

\subsection{Three-Stage Progressive Learning}
The training process of CombatVLA adheres to a three-step progressive learning paradigm, enabling the model to gradually master combat strategies. Initially, the model undergoes coarse-grained video-level training (stage1), followed by fine-grained frame-level training (stage2), and finally truncation strategy training (stage3). During the training process, we froze the parameters of the vision encoder and fine-tuned the parameters of the language model.
%By applying a curriculum learning approach, training a machine learning model in an ordered fashion from simple to complex samples can enhance its performance without increasing computational costs. In this work, we will fine-tune the model in the order from coarse-grained videos to fine-grained keyframes, as shown in Fig.{xx}.During the training process, we froze the parameters of the vision encoder and fine-tuned the parameters of the language model.

\noindent\textbf{Stage1: Coarse-Grained Video-AoT Tuning.}
The goal of this training stage is to help the model understand the combat environment, make learning easier, and stabilize training. Regarding the training data, Video-AoT, each video consists of $n$ frames, with a frame rate set to $m$ frames per second. We arrange the actions corresponding to each frame in chronological order, thereby generating video-AoT data pairs. Notably, actions are not precisely timed with frames, so the model must infer actions from the overall visual content rather than exact timing. This strategy enables our model to consider all possible actions and gain an initial understanding of the combat paradigm.
%To reduce the initial learning difficulty and stabilize the training dynamics, we first segment and connect video frames in the dataset, with each segment containing $n$ frames, and set the frame rate to $m$ frames per second. Simultaneously, we arrange the actions corresponding to each frame in chronological order, thereby generating video-AoT data pairs. In these data pairs, actions are not precisely aligned with the video frames in time, meaning that for a given segment of video, the model needs to judge and execute the corresponding actions based on the overall visual content, without focusing on the specific timing of each action in the frames. Therefore, we help the model to consider all possible actions comprehensively on the basis of video segments, thereby gaining an initial understanding of the combat paradigm in a global temporal space.

\noindent\textbf{Stage2: Fine-Grained Frames-AoT Tuning.}
In 3D combat games, precise second-level reaction time is crucial, requiring the model to quickly understand environment and make fast decisions. In this stage, we create action-frame aligned data pairs, called Frames-AoT, tracing back $k$ frames from the current action's timestamp. For example, if $k$ frames show the enemy preparing to attack, the model might decide to dodge. This strategy helps our model understand the sequence and logic of combat scenarios.
%In 3D combat gameplay, the demand for reaction speed is precise to the second-level level, requiring the model to extract visual information from a minimal number of frames and make rapid action decisions. In this process, we select the timestamp of each action-corresponding frame as the starting point and retrieve a total of $k$ frames leading up to it. These frames, along with the corresponding actions, form a frames-AoT data pair, which enables the model to learn the correct action to take when presented with this sequence of $k$ frames. For example, if the sequence of $k$ frames depicts an enemy's attack wind-up, the appropriate action might be to dodge. Through this fine-grained tuning, the model can focus more on the temporal sequence details, gaining a deeper understanding of the rapid transitions within the local temporal space of combat scenarios.

\noindent\textbf{Stage3: Fine-Grained Frames-Truncated-AoT Tuning.}
The inference time of LLM/VLM-based models is proportional to token length due to the next token prediction requirement. Therefore, we designed a truncation strategy to mitigate the time increase associated with the introduction of AoT. As shown in Fig.\ref{fig:AoT}, we reorganized the AoT data by introducing a special token $ \langle \text{TRUNC} \rangle $. During real-time operations, any response following $ \langle \text{TRUNC} \rangle $ will be truncated. This strategy allows our model to retain the benefits provided by AoT while accelerating the inference process.
%Due to the next-token prediction characteristic of autoregressive models, the total inference time of LLMs/VLMs is proportional to the number of tokens generated. In this work, we introduce the AoT data structure, which contains rich semantic explanations. This naturally increases the total number of tokens that need to be generated during the inference phase, thereby extending the inference time. However, during inference, we actually only need the model's output to be actions, without lengthy explanations. Therefore, we reorganize the AoT data as shown in Fig.{xx}. In this structure, actions are placed in the first half, with the related explanations in the latter half, separated by a special token $<trunc>$. This arrangement forms frames-truncated-AoT data pairs, allowing us to discard the explanation part after $<trunc>$ during the inference stage.
%We further utilize these truncated-AoT data to fine-tune the model, with the goal of maintaining the performance gained from the previous two stages of AoT training while converting the model's output format to truncated AoT, thus supporting faster inference. Through this strategy, we effectively optimize the model's inference efficiency.

\paragraph{Adaptive Action-Weighted Loss.}
% 公式
% To address the challenge of imbalanced action distributions in training data (particularly for rare but critical actions like health recovery), we propose an adaptive action-weighted loss that enhances cross-modal alignment between visual inputs and action semantics while improving sensitivity to low-frequency actions. 
% Given visual input  and action-of-thought sequence , the vision encoder and projector generate a vision embedding sequence \( \tilde{S}_v = [v_1, \ldots, v_m, v_{\text{EOS}}] \), where \( v_{\text{EOS}} \) captures the global visual representation via an end-of-sequence token. Similarly, the action sequence is embedded as \( \tilde{S}_a = [a_1, \ldots, a_n, a_{\text{EOS}}] \), with \( a_{\text{EOS}} \) encoding global action semantics. These sequences are processed by the LLM to produce final embeddings \( \hat{v}_{\text{EOS}} \) and \( \hat{a}_{\text{EOS}} \), which serve as global cross-modal representations.  

Our CombatVLA is trained with three losses—language modeling $\mathcal{L}_{lang}$, action alignment $\mathcal{L}_{align}$, and modality contrastive $\mathcal{L}_{con}$—to address action distribution imbalance.
%To address the issue of imbalanced action distribution in the dataset (such as low-frequency actions like health restoration), we propose an adaptive action-weighted loss to enhance the model's sensitivity to rare actions.

Firstly, to better capture the correspondence between vision and action, following \citet{he2025analyzing}, we introduce a contrastive loss. 
Specifically, given an input visual image $V$ and action-of-thought data $A$, the visual [EOS] and the final [EOS] of the LLM output serve as the local representations of $V$ and $A$, respectively. Please refer to Fig.\ref{fig:framework}(d) for a quick understanding.
%Specifically, given an input visual image $V$ and action-of-thought data $A$, we use a vision encoder and projector to transform $V$ into a vision embedding sequence \( S_v = [v_1, v_2, ..., v_m] \). Following \citet{he2025analyzing}, we input an [EOS] token into the embedding layer of the LLM to obtain a better global representation, denoted as \( v_{EOS} \). We append this to \( S_v \) to obtain \( \tilde{S_v} = [v_1, v_2, ..., v_m, v_{EOS}] \). Similarly, we obtain the AoT embedding sequence \( \tilde{S_{a}} = [a_1, a_2, ..., a_n, a_{EOS}] \). Then, we input \( \tilde{S_v} \) and \( \tilde{S_{a}} \) into the LLM, and the embeddings of the last predicted token \( \hat{v}_{EOS} \) and \( \hat{a}_{EOS} \) are taken as the global representations of V and A, respectively.
Subsequently, based on whether the actions output by the model \( A_o \) \textbf{match} the actions in the label \( A_l \), we adjust the distance between the embeddings \( \hat{v}_{EOS} \) and \( \hat{a}_{EOS} \), either bringing them closer or pushing them apart.

A subsequent question is: how do we determine whether \( A_o \) and \( A_l \) match? We introduce a priority-aware matching criterion based on a predefined action sequence \( P = [c_0, \ldots, c_{k-1}] \), where \( c_0 \) denotes the highest-priority action category (ranked by functional importance and occurrence frequency). The matching function \( \mathcal{M}(A_l, A_o) \) evaluates whether the highest-priority action category $c^*$ in $A_l$ (determined by $P$) exists in $A_o$:
\begin{equation}
\mathcal{M}(A_l, A_o) = \begin{cases}
1 & \text{if } \mathop{\arg\max}\limits_{c^* \in P}\mathbb{I}(c^* \in A_l) \in A_o \\
0 & \text{if } \mathop{\arg\max}\limits_{c^* \in P}\mathbb{I}(c^* \in A_l) \notin A_o
\end{cases}
\end{equation}

% The loss function dynamically adapts to the matching result: for matched pairs (\( \mathcal{M}=1 \)), we minimize the cosine distance between \( \hat{v}_{\text{EOS}} \) and \( \hat{a}_{\text{EOS}} \) via \( \mathcal{L}^{pull}_{\text{cross}} = 1 - \cos(\hat{v}_{\text{EOS}}, \hat{a}_{\text{EOS}}) \); for mismatched pairs (\( \mathcal{M}=0 \)), we maximize their separation using \( \mathcal{L}^{push}_{\text{cross}} = \cos(\hat{v}_{\text{EOS}}, \hat{a}_{\text{EOS}}) - 1 \) while enforcing action prediction accuracy through an alignment loss \( \mathcal{L}_{\text{align}} = -\sum \log p(c^*) \). The composite action loss is defined as:  

The loss function dynamically adapts to the matching result: for matched pairs (\( \mathcal{M}=1 \)), we minimize the cosine distance between \( \hat{v}_{\text{EOS}} \) and \( \hat{a}_{\text{EOS}} \) via:
\begin{equation}
 \mathcal{L}^{pull}_{\text{con}} = 1 - \cos(\hat{v}_{\text{EOS}}, \hat{a}_{\text{EOS}}) 
\end{equation}
and for mismatched pairs (\( \mathcal{M}=0 \)), we maximize their separation using \( \mathcal{L}^{push}_{\text{con}} = -\mathcal{L}^{pull}_{\text{con}} \) while enforcing action prediction accuracy through an alignment loss \( \mathcal{L}_{\text{align}} = -\sum \log p(c^*) \). The composite action loss is defined as: 

\begin{equation}
\mathcal{L}_{\text{act}} = \begin{cases} 
\mathcal{L}^{pull}_{\text{con}} & \text{if } \mathcal{M}=1 \\ 
\mathcal{L}^{push}_{\text{con}} + \mathcal{L}_{\text{align}} & \text{if } \mathcal{M}=0 
\end{cases}.
\end{equation}

The final objective combines language modeling \( \mathcal{L}_{\text{lang}} \):

\begin{equation}
\mathcal{L} = \mathcal{L}_{\text{lang}} + \alpha \cdot \mathcal{L}_{\text{act}},
\end{equation}
where \( \alpha \) is derived from action priorities using exponential weights \( \alpha_i = 2^{(k-i-1)} \) normalized to the range [0.1, 1.0]. Here, \( k \) represents the length of \( P \) or the number of action classes, and \( i \) is the index of action \( c^* \) within \( P \). This formulation prioritizes rare critical actions, ensuring balanced learning and strong vision-action alignment, despite imbalanced action categories.

\begin{table}[t]
\centering
\caption{Task definitions in \textit{Black Myth: Wukong} (BMW) and \textit{Sekiro: Shadows Die Twice} (SSDT).}
\vspace{-0.3cm}
\resizebox{0.45\textwidth}{!}{
\begin{tabular}{c|c|c|c|c}
\toprule
\textbf{Game} & \textbf{Task ID} & \textbf{Description} & \textbf{Diffuculty} & \textbf{Zero-Shot} \\ \hline
\multirow{10}{*}{BMW} & 1 &	Defeat WolfScout & Easy & \cmark \\ \cline{2-5}
 & 2 & Defeat WolfStalwart & Easy & \cmark \\ \cline{2-5}
 & 3 &	Defeat WolfSwornsword & Easy & \cmark \\ \cline{2-5}
 & 4 & Defeat WolfSoldier & Easy & \cmark \\ \cline{2-5}
 & 5 & Defeat Croaky & Easy & \cmark \\ \cline{2-5}
 & 6 & Defeat Crow Diviner & Middle & \cmark \\ \cline{2-5}
 & 7 & Defeat Bandit Chief & Middle & \cmark \\ \cline{2-5}
 & 8 & Defeat Bullguard & Hard & \cmark \\ \cline{2-5}
 & 9 & Defeat Wandering Wight & Very Hard & \xmark \\ \cline{2-5}
 & 10 & Defeat Guangzhi & Very Hard & \xmark \\ \hline
\multirow{3}{*}{SSDT} & 11 & Defeat Katana & Easy & \cmark \\ \cline{2-5}
 & 12 & Defeat Hassou Stance & Middle & \cmark \\ \cline{2-5}
 & 13 & Defeat Shigenori Yamauchi & Hard & \cmark \\
\bottomrule
\end{tabular}} \label{tab:task}
\vspace{-0.5cm}
\end{table}

\subsection{Action Execution Framework}
\noindent\textbf{VLA-based Agent Framework.}
To support VLMs in playing computer games like humans, we developed a lightweight and fast action execution agent. For instance, our fine-tuned VLM is similar to the human brain, responsible for reasoning and decision-making, while the framework is akin to human eyes and hands, responsible for observation and execution.
In real-time PC gameplay, the input of the action execution agent is the real-time game video footage captured, and the output is actions based on mouse and keyboard operations. Specifically, we perform frame sampling on the captured real-time game footage at over 60 FPS, removing redundant visual information to reduce the computational pressure on VLMs during inference. Finally, the model's output adopts a truncated inference strategy to extract useful action information for execution.

% \paragraph{Adaptive-Frame Sampling.}
\noindent\textbf{Truncated Inference and Execution.}
During inference, we monitor each new output token and stop when we see the $ \langle \text{TRUNC} \rangle $ token, converting prior tokens into actions. This strategy accelerates the inference speed. Subsequently, we translate the actions into Python code using the ``pyautogui'' library to automate mouse and keyboard controls, enabling the game character to execute combat tasks.
% Since we used frames-truncated-AoT tuning in the final stages of the curriculum-learning action tuning, the model's inference outputs will follow the truncated AoT format. During inference, we utilize the streaming response, continuously monitoring each new output token. When the $<trunc>$ token is encountered, it triggers a truncation, retaining all preceding tokens, converting them into actions, and stopping further inference. This approach can significantly increase the model's inference speed. We then convert the resulting actions into Python code format and use the pyautogui library to automatically control the mouse and keyboard, allowing the game character to perform combat tasks.
% % Stream response and truncated condition
% % pyautogui for execution

%% file: sec/4_experiment.tex
\begin{table*}[t]
\centering
\caption{Performance comparison of closed source and open source LVLMs on the combat understanding benchmark and general benchmark. The highest scores among models in each metric are highlighted in \textbf{FirstBest}.}
\label{tab:model-performance}
\vspace{-0.2cm}
\begin{tabular}{lcccc|ccc}
\toprule
\multirow{2}{*}{\textbf{Model}} & \multicolumn{4}{c}{\textbf{Combat Understanding}} & \multicolumn{3}{c}{\textbf{General Benchmark}} \\ \cline{2-8}
                                & Gathering   & Comprehension  & Reasoning  & Avg.  & MME        & VideoMME        & OCRBench        \\
\midrule
\multicolumn{8}{c}{\textit{Closed-Source Large Vision Language Models}} \\
\midrule
GPT-4o-0513                     & 58.06&	\textbf{66.67}&	47.14&	57.29& 2328          & 71.9               & 736               \\
GPT-4o-mini-0718                & 59.44&	66.18&	42.57&	56.06& 2003          & 64.8               & 785               \\
GPT-4-vision-preview            & 52.78&	53.92&	43.71&	50.14& 1926          & 59.9               & 645               \\
Gemini-2.0-flash                & 58.61&	64.22&	50.86&	\textbf{57.90} & --          & --               & --               \\
Gemini-1.5-pro                  & \textbf{64.44}&	62.75&	41.71&	56.30& 2110          & 75.0               & 754               \\
Claude35-sonnet & 53.89&	57.35&	\textbf{55.43}&	55.56& 1920          & 60.0               & 788               \\
\midrule
\multicolumn{8}{c}{\textit{Open-Source Large Vision Language Models}} \\
\midrule
% LLaVA-NeXT-Video-7B    & 62.50 &	44.61 &	14.29 &	40.47 & -          &  -               & -               \\
LLaVA-1.5-7B & 50.56 & 60.29 & 42.86 & 51.24 & 1510 & -- & -- \\
InternVL2.5-4B    & 53.89 &	48.04 &	43.71 &	48.55 & 2337          &  62.3               & 828               \\
% qwen-vl-max                     & 68.61&	42.65&	41.43&	50.90& 0          & 0               & 0               \\
Qwen2-VL-7B            & 55.28&	59.80&	43.14&	52.74& 2326          &  63.3               & 866               \\
Qwen2-VL-2B            & 53.33&	46.57&	42.86&	47.59& 1872          &  55.6               & 809               \\
Qwen2.5-VL-7B                   & 45.56&	52.94&	50.57&	49.69& 2347          & 65.1               & 864               \\
Qwen2.5-VL-3B                   & 53.61&	56.86&	57.14&	55.87& 2157          & 61.5               & 797               \\
\midrule
% CombatVLA\_s1 (Ours)                   & 53.89&	57.35&	60.57&	57.27     & --          & --               & --               \\
% CombatVLA\_s2 (Ours)                   & 59.17&	60.29&	62.86&	60.77     & --          & --               & --               \\
\rowcolor[gray]{0.9}\textbf{CombatVLA-3B (Ours)}                   & \textbf{60.83}           & \textbf{60.29}              & \textbf{69.71}          & \textbf{63.61}     & 2141          & 58.7               & 741              \\
\bottomrule
\end{tabular}
\vspace{-0.2cm}
\end{table*}

\section{Experiments}
\subsection{Implementation Details}
\noindent\textbf{Dataset.}
% task defination
Following VARP~\cite{chen2024varp}, we employ the two games,  ``Black Myth: Wukong(BMW)" and ``Sekiro: Shadows Die Twice(SSDT)'', as experimental platforms. The annotators defined 13 combat tasks based on their difficulty levels, categorizing them into four distinct levels: easy, medium, hard, and very hard, as shown in Tab.~\ref{tab:task}.

% dataset
We collected training data from task 9 and 10 of ``Black Myth: Wukong'' using our action tracker. After manual selection, we organized approximately 25k game screenshots and 5k high-quality AoTs, split into 95\% for training and 5\% for validation. The AoTs includes 10 actions: ``wsad'' for movement, ``shift'' for quick moves, ``space'' for dodging, ``r'' for healing, ``1'' for immobilization, ``left mouse button'' for light attacks, and hold ``right mouse button'' for heavy attacks. Notably, these actions can be combined.
% We collected extensive data using the action tracker on the two very hard tasks, task 9 and task 10 of ``Black Myth: Wukong''. After meticulous manual selection, we organized approximately 25k game-screenshot images and 5k high-quality AoT to construct the combat game dataset, of which 95\% is designated as the training set and 5\% as the validation set. 
% The AoT includes 10 types of actions: 'wsad' for movement (up, down, left, right), 'shift' for quick movement, 'space' for dodging, 'r' for health recovery, '1' for using an immobilization skill, 'left mouse button' for light attacks, and long-press 'right mouse button' for heavy attacks. These actions can be combined.

\noindent\textbf{Benchmarks.}
% benchmark
We evaluated baselines using the combat understanding benchmark (CUBench), general benchmarks (i.e., MME~\cite{fu2024mme}, VideoMME~\cite{fu2024videomme}, OCRBench~\cite{Liu2024ocrbench}), and practical tests. 
%To accurately assess the model’s performance in this domain, we also created a combat understanding benchmark (CUBench), which consists of three aspects: information gathering, combat comprehension, and action reasoning, with levels ranging from low to high. We further evaluated the model's capabilities on general benchmarks, including MME, VideoMME, and OCRBench.
% task-level practical tests
In task-level practical tests, the action execution framework controls the PC to engage in real combat. All baselines are tested 10 times per task, with success marked by defeating the enemy and failure by being defeated. We calculate success rates from these attempts and record average inference time. 
Notably, our CombatVLA, fine-tuned only on very hard tasks (tasks 9 and 10), uses tasks 1-8 (same game, different tasks) and 11-13 (different game, different tasks) as zero-shot tests for assessing generalization. For fair comparison, we made adaptation improvements to Cradle framework~\cite{tan2024cradle} for the two games.
% Additionally, we employed the action execution agent to enable CombatVLA to operate a PC and play ARPG combat tasks like a human, thereby completing task-level practical tests. Specifically, CombatVLA will be tested 10 times for each task, considering a task completed successfully if the enemy is defeated and a failure if the player is defeated. We calculate the success rate from these 10 trials and record the average inference time per model call. Since CombatVLA was fine-tuned only on data from the very hard tasks, task 9 and task 10, tasks 1-8 (same game, different tasks) and tasks 11-13 (different game, different tasks) will serve as zero-shot tests for CombatVLA, thus more accurately testing the model’s generalization ability in executing combat tasks. For a fair comparison, we made adaptation improvements to Cradle for the two games.

\noindent\textbf{Baselines.} 
%To compare with existing VLM/VLA models, 
We selected closed-source models such as GPT-4o, GPT-4o-mini, GPT-4-vision-preview\footnote{\url{https://openai.com/index/}}, Gemini-2.0-flash\footnote{\url{https://deepmind.google/technologies/gemini/flash/}}, Gemini-1.5-pro \cite{team2024gemini}, and Claude3.5-Sonnet\footnote{\url{https://www.anthropic.com/news/claude-3-5-sonnet}}, as well as open-source models of similar sizes like LLaVA-1.5-7B \cite{liu2024llava_1_5}, InternVL2.5-4B \cite{chen2025internvl2_5}, Qwen2-VL-7B/2B \cite{wang2024qwen2} and Qwen2.5-VL-7B/3B\footnote{\url{https://help.aliyun.com/zh/model-studio/developer-reference/use-qwen-by-calling-api}}, as baselines for the CUBench and general benchmarks. 
Cradle \cite{tan2024cradle}, VARP \cite{chen2024varp} and ten human players were chosen as baselines for task-level practical tests.

% benchmark
% action-level acc: alot of baselines..
% task-level acc：Cradle，VARP
% action frequency
% total inference time
% 要不要测通用VQA能力？VideoMME, MMBench-Video, LV-Bench, MME-RealWorld

\begin{figure*}[h]
\vspace{-0.8cm}
\centering
\includegraphics[width=1.0\textwidth]{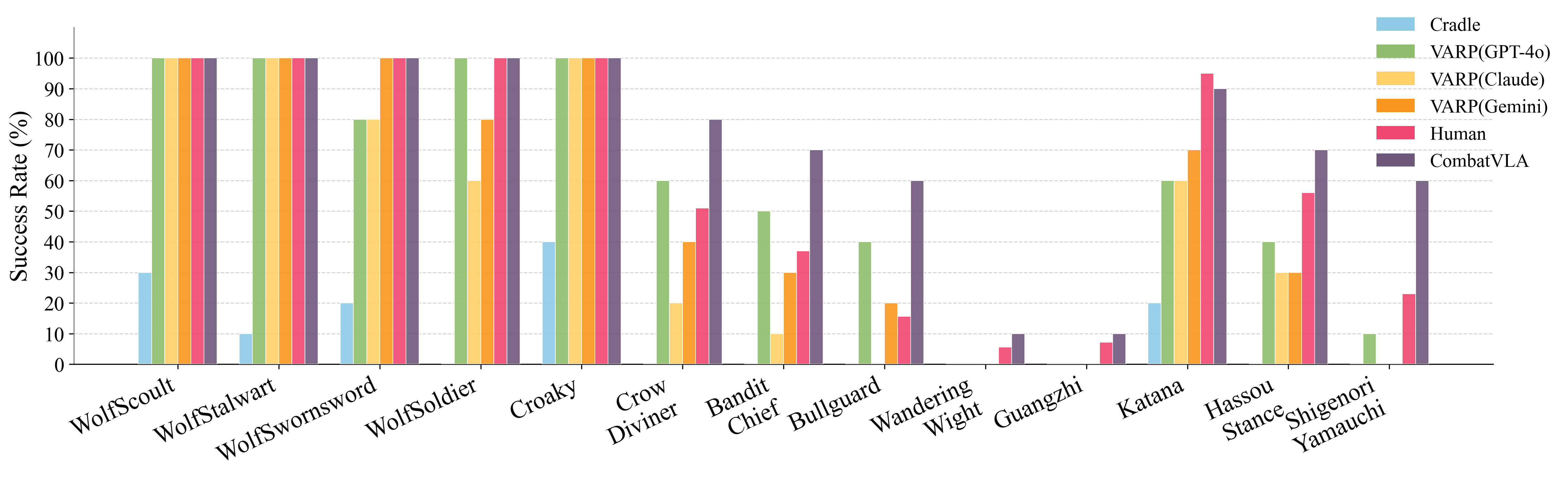}
\vspace{-0.8cm}
\caption{Comparison of task-level practical tests. Our CombatVLA not only outperforms all VLM-based agents (i.e., Cradle and VARP) but also has a higher task success rate than human players.}
\label{fig:task_level}
\vspace{-0.4cm}
\end{figure*}

\begin{table}[]
\centering
\caption{Comparison of latency and model calls per inference.}
\label{tab:speed}
\vspace{-0.2cm}
\begin{tabular}{lrr}
\toprule
               Method  & Latency(s)$\downarrow$ & Model Calls$\downarrow$ \\
\midrule
Cradle~\cite{tan2024cradle}           & 61.68               & 5         \\
VARP~\cite{chen2024varp}             & 90.23               & 10         \\
\rowcolor[gray]{0.9}\textbf{CombatVLA (Ours)}        & \textbf{1.85}            & \textbf{1}         \\
\bottomrule
\end{tabular}
\vspace{-0.3cm}
\end{table}

\noindent\textbf{Training Settings.}
Our backbone leverages Qwen2.5-VL-3B with a full-parameter supervised fine-tuning(SFT) on our AoT dataset. We have configured a learning rate of 1e-5, with a batch size of 1, and set the temperature to 0.7. During the coarse-grained video-AoT tuning phase, we use $n=20$ and $m=10$, training for 3 epochs. In the subsequent fine-grained frames-AoT tuning phase, we set $k=4$ and train for 1 epoch. The finally stage3 phase is carried out over 3 epochs. All training and benchmark evaluations were conducted on 4 NVIDIA H20 GPUs. Task-level pratical tests were conducted on a NVIDIA RTX 4090 GPU.

\subsection{Main Results}
\noindent\textbf{Combat Understanding Evaluation.}
We evaluated CombatVLA and the baselines on CUBench and general benchmarks, with experimental results shown in Tab.\ref{tab:model-performance}. On CUBench, our CombatVLA achieved the highest average score of 63.61, surpassing the second-highest Gemini-2.0-flash by 5.71 points. Compared to the original backbone, Qwen2.5-VL-3B, it improved by 7.74 points. This indicates that our method significantly enhances the model's capability in combat understanding. Specifically, while ours did not perform best in low-level abilities such as information gathering and combat comprehension, it still demonstrated considerable competitiveness. However, in high-level action reasoning, ours outperformed the second-highest Claude35-sonnet by 14.28 points, thanks to the action-of-thought data enhancing the model's reasoning capability.
% general

\noindent\textbf{General Benchmark Evaluation.}
MME, VideoMME, and OCRBench are representative benchmarks for image, video, and rich text, respectively. They are used to evaluate the impact of AoT training on general capabilities. As shown in Tab.\ref{tab:model-performance}, despite being trained on specific tasks, our CombatVLA maintains comparable performance to the backbone model Qwen2.5-VL-3B. This further confirms the robustness of our approach.

\noindent\textbf{Task-Level Practical Evaluation.}
We integrated CombatVLA into the action execution agent to play the game like a human, automatically carrying out combat tasks. Due to unavoidable inference delays with the model, the agent pauses the game during model inference and resumes the game when executing actions. To ensure the fairness and feasibility of the experiment, we set all methods to have the game character's attack attribute at 100 and defense attribute at 600 when testing task 9 and task 10. The key settings for SSDT and BMW are consistent, such as the block action in SSDT and the dodge action in BMW, both of which are assigned to the ``space'' key.

The experimental results are shown in Fig.~\ref{fig:task_level}. The following observations can be made: 1) Although we made adaptation improvements for Cradle, its reasoning heavily relies on explicit text prompts within the game screenshot. Since combat tasks require the ability to extract implicit visual information, Cradle performed the worst across all tasks. 2) VARP, despite extensive engineering adaptations for the BMW game, performed poorly on BMW's hard and very hard difficulty tasks. Moreover, VARP's success rate in the SSDT game significantly decreased, indicating low generalization capability. 
3) Our CombatVLA, apart from being comparable to humans on some easy tasks, surpassed baselines on the other tasks, especially the hard and very hard tasks. Zero-shot tests on tasks 1 to 8 of the same game, and tasks 11 to 13 of a different game, also demonstrated CombatVLA's strong generalization ability.

\noindent\textbf{Inference Latency.}
We also compared the average latency per inference (i.e., the process of inputting visual information and outputting actions) and the number of model calls with the baselines, as shown in Tab.~\ref{tab:speed}. Our model requires only 1.8 seconds of delay and one model call, compared to VARP, representing a speed increase of 50 times and reducing model call costs to $\frac{1}{10}$ of VARP's. For more detailed information and visualizations, please refer to our supplementary materials and demo videos.

% success rate
% latency per inference
% model calls per inference

\begin{table}[t]
\centering
\caption{Ablation study of progressive learning.}
\label{tab:training}
\vspace{-0.2cm}
\resizebox{0.5\textwidth}{!}{
\begin{tabular}{cccccc}
\toprule
Training & Gathering & Comprehension & Reasoning & Avg. & Time(s)$\downarrow$ \\
\midrule
Stage1           & 53.89  & 57.35 & 60.57 & 57.27 &   3.73      \\
Stage2           & 59.17  & \textbf{62.25} & 62.86 & 61.43 &   3.73      \\
Stage3           & \textbf{60.83}  & 60.29 & \textbf{69.71} & \textbf{63.61} & \textbf{1.85}         \\
\bottomrule
\end{tabular}
}
\end{table}

\begin{table}[t]
\centering
\caption{Ablation study of adaptive action-weighted loss.}
\label{tab:loss}
\vspace{-0.2cm}
\resizebox{0.5\textwidth}{!}{
\begin{tabular}{lcccc}
\toprule
Loss Setting & Gathering & Comprehension & Reasoning & Avg. \\
\midrule
Stage3           & 60.83  & \textbf{60.29} & \textbf{69.71} & \textbf{63.61}         \\
w/o $\mathcal{L}_{con}$           & \textbf{62.78}  & 58.82 & 63.14 & 61.58         \\
w/o $\mathcal{L}_{align}$           & 61.39  & 59.80 & 63.71 & 61.64         \\
\bottomrule
\end{tabular}
}
\end{table}

\begin{figure*}[h]
\vspace{-0.8cm}
\centering
\includegraphics[width=0.96\textwidth]{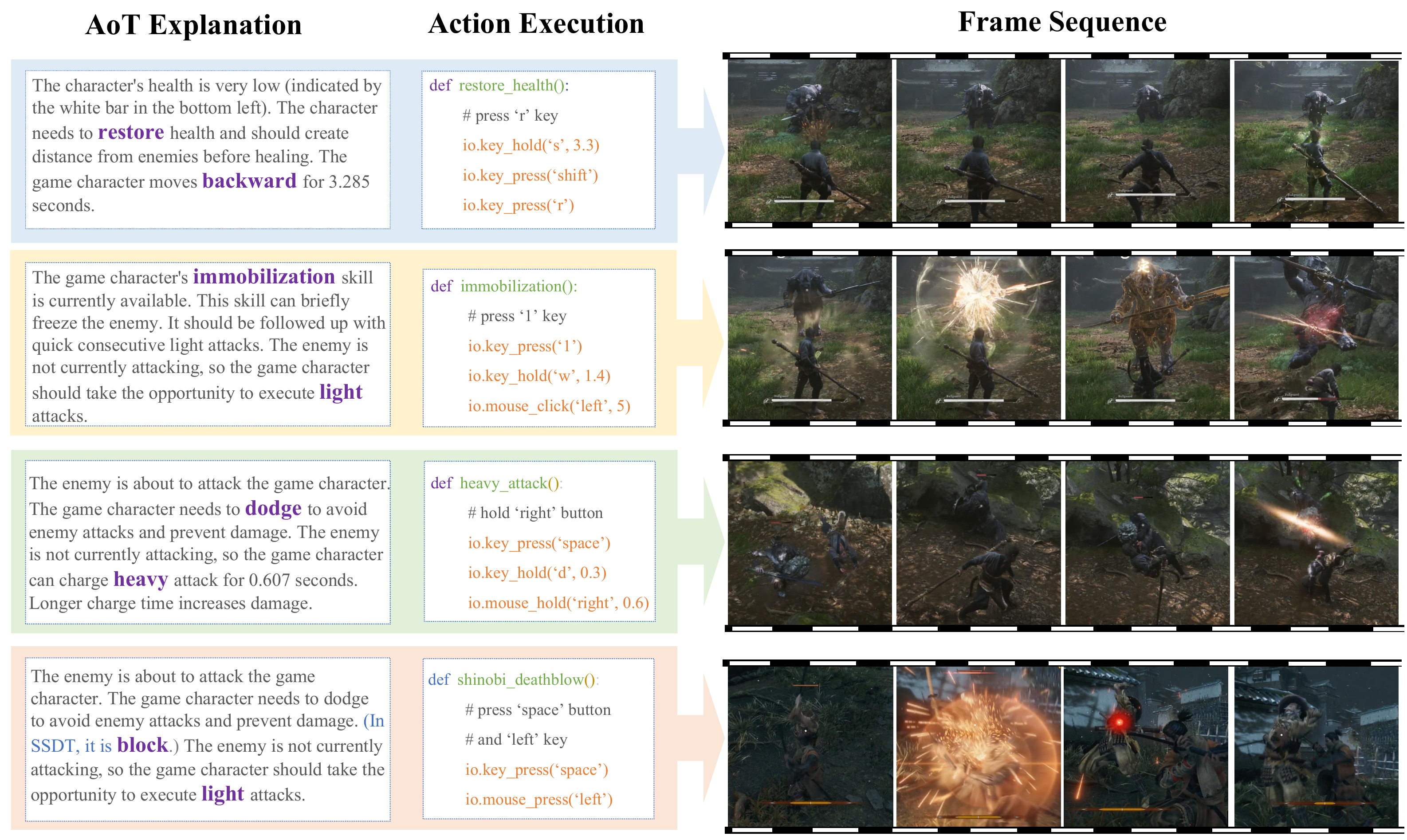}
\vspace{-0.2cm}
\caption{Visualizations of some representative cases involving BMW and SSDT are provided, along with explanations of the model outputs, the corresponding action execution codes, and the frame-by-frame sequences.}
\label{fig:vis}
\vspace{-0.4cm}
\end{figure*}

\subsection{Ablation Studies}
\noindent\textbf{Ablation of Progressive Learning.}
We evaluated the different stages of progressive learning, as well as the experimental results of \( \mathcal{L}_{\text{con}} \) and \( \mathcal{L}_{\text{align}} \) on CUBench. As shown in Tab.~\ref{tab:training}, 
%\( S1 \) indicates the model is trained only up to stage 1, \( Stage2 \) indicates the model is trained up to stage 2, and \( Stage3 \) represents our full pipeline model. 
Stage3 represents our full CombatVLA model. 
The following observations can be made: 1) \( Stage1 \) can only learn coarse-grained actions from the video, but the actions are not aligned with specific frames in the video, so its performance is the worst. 2) Additionally, Tab.~\ref{tab:training} reports the average inference time for invoking the model once, and due to the truncation mechanism, the speed of \( Stage3 \) is about 2 times that of \( Stage2 \).

\noindent\textbf{Ablation of Adaptive Loss.}
As progressive learning proceeds, the model's performance in gathering and reasoning gradually improves. Due to the introduction of the $ \langle \text{TRUNC} \rangle $ token in \( Stage3 \), which places the action's explanation after the action and truncates it, the model cannot access the semantic information when generating actions, thus somewhat impairing the model's understanding performance. However, as shown in Tab.~\ref{tab:loss}, the introduction of \( \mathcal{L}_{\text{con}} \) and \( \mathcal{L}_{\text{align}} \) enhanced the model's reasoning performance, reaching 69.71, which is 6.85 points higher than \( Stage2 \), with an average score increase of 2.18 points.

\subsection{Qualitative Visualization}
We demonstrated some representative cases from the task-level practical tests. We reported the AoT explanations inferred by CombatVLA, the actions parsed into Python code, and the sequence of frames after executing the actions, as shown in Fig.~\ref{fig:vis}. The first three rows are belong to BMW, and the fourth is SSDT. 
We have the following observations:
\begin{itemize}[leftmargin=*]
 \item{In the first row, CombatVLA detected its own low health and decided to immediately use the restore health action. Therefore, it first moved the game character backward to a safe position and then pressed the ``r'' key to increase its health.}
 \item{In the second row, CombatVLA detected that its immobilizing skill was in an available state, so it pressed the ``1'’ key to immobilize the enemy, then immediately launched a series of continuous attacks, depleting the enemy’s health significantly. }
 \item{The third row shows how our model effectively dodged the enemy’s attack and then used a heavy attack with a wind-up at an opportune moment.}
 \item{In the fourth row, CombatVLA first used the block action to withstand an enemy attack, then executed a light attack to perform a shinobi deathblow, killing the enemy in one hit. }
\end{itemize}
Overall, CombatVLA demonstrates a strong ability to understand combat tasks and can effectively reason out for various complex situations with the help of advanced semantic information from AoT explanations.

%% file: sec/5_conclusion.tex
\section{Conclusion}
In this paper, we aim to address the issue that VLMs or VLAs lack second-level response times, high-resolution perception, and tactical reasoning in 3D action role-playing games. Specifically, we introduce CombatVLA, a 3B model trained on AoT sequences with the constraint of action alignment loss and modality contrastive loss.  Thereafter, CombatVLA seamlessly integrates into an action execution framework, allowing efficient inference through our truncated AoT strategy.  Experimental results demonstrate that our CombatVLA not only surpasses all existing models in combat understanding benchmarks, while maintaining generalization capability, but also achieves a 50-fold speedup in real-time combat scenarios.
In the future, we will further enhance model's understanding of game scenarios, thereby expanding its application to more games.

% \section{Limitations}
% We must also candidly acknowledge some limitations in our research, specifically: 1) Task Definitions: As VLM- and VLA-based agents are still evolving, the current task definitions are somewhat simplistic. 2) Game Scenarios: Our research has only been tested within the BMW and SSDT game and has not yet been extended to other scenarios. 3) Model Capabilities: As shown in the benchmark evaluation section, there is still room for improvement in existing VLMs and VLAs. 

%% file: sec/X_supple.tex
\clearpage
\setcounter{page}{1}
\maketitlesupplementary

% \section{Appendix}
\section{Overview}
\begin{itemize}
    \item Limitations (\S \ref{sec:limit})
    \item More Details (\S \ref{sec:details}) 
    % \item Details of Data Annotation (\S \ref{sec:annotation})
    % \item Details of Prompts (\S \ref{sec:prompt})
    % \item Details of CUBench Benchmark (\S \ref{sec:dataset})
    % \item Details of Action Tracker (\S \ref{sec:tracker})
    % \item Details of Action-of-Thought Explanation (\S \ref{sec:aot})
    \item Additional Qualitative Visualization (\S \ref{sec:case})
    \item Task Defination (\S \ref{sec:task})
    \item Token Length of AoT (\S \ref{sec:length})
    \item VQA Reasoning Case of CUBench (\S \ref{sec:cubench_case})
    \item Demo Video (\S \ref{sec:video})
\end{itemize}

\section{Limitations} \label{sec:limit}
We must also candidly acknowledge some limitations in our research, specifically: 1) Task Definitions: As VLM- and VLA-based agents are still evolving, the current task definitions are somewhat simplistic. 2) Game Scenarios: Our research has only been tested within the BMW and SSDT game and has not yet been extended to other scenarios. 3) Model Capabilities: As shown in the benchmark evaluation section, there is still room for improvement in existing VLMs and VLAs.

\section{More Details} \label{sec:details}

\subsection{Details of Data Annotation} \label{sec:annotation}
The game annotation team comprises six individuals, each of whom has completed all levels of the game. Over a two-week period, their gameplay data was recorded using our action tracker. After filtering out abnormal samples with insufficient action density, we cleaned 200 hours of recordings, including video, mouse, and keyboard inputs.

The data annotation team consists of ten members, each with at least a bachelor's degree and gaming experience. They are responsible for annotating the benchmark data and the formatted AoT data generated by GPT-4o. All QA pairs are annotated by the this team and cross-validated to ensure high quality. This validation process ensures that only data passing all quality checks is retained.

Ultimately, we compiled 914 data fragments for our CUBench, 25,000 game screenshots with a resolution of $1008 \times 560$, and 5,000 high-quality AoTs.

\begin{figure*}[h]
\vspace{-0.8cm}
\centering
\includegraphics[width=1.0\textwidth]{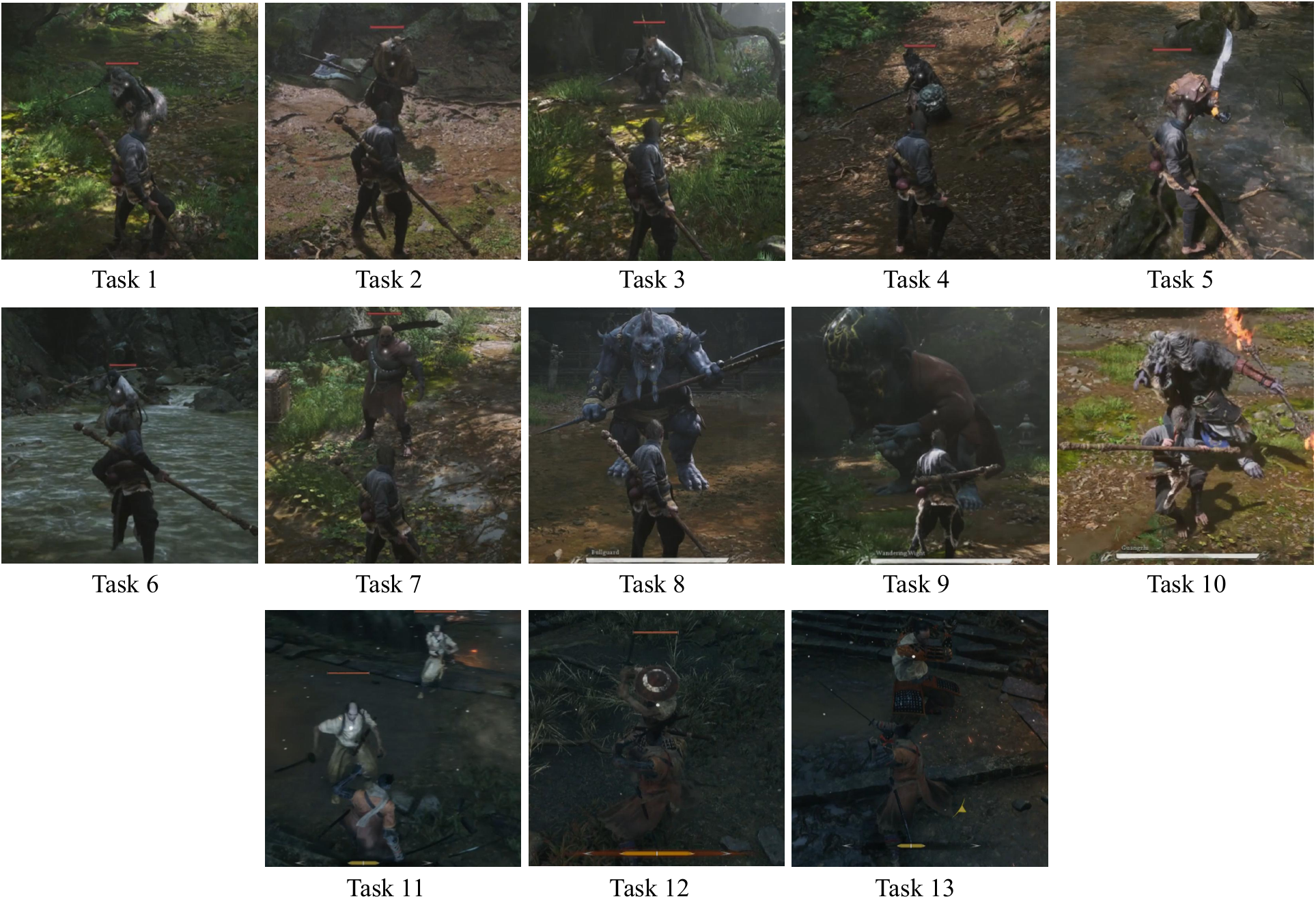}
% \vspace{-0.8cm}
\caption{The visualization of 13 defined tasks.}
\label{fig:task}
% \vspace{-0.4cm}
\end{figure*}

\subsection{Details of Prompts} \label{sec:prompt}
The prompts for QA pair generation of combat understanding benchmark(CUBench) are as follows,
\vspace{-0.6cm}
\begin{center}
\fcolorbox{black}{gray!10}{\parbox{1\linewidth}{
\textbf{Prompts of Benchmark Collection} \\
\\
\textbf{----- Gathering -----} \\
\textbf{Gathering enemy health} \\
Select the best answer to the following single-choice question based on the game-screenshot image. Respond with only the letter (Yes or No) of the correct option. Is the enemy's health high in the game?  Yes/No. The best answer is: \\
\textbf{Gathering own health} \\
Select the best answer to the following single-choice question based on the game-screenshot image. Respond with only the letter (Yes or No) of the correct option.Is the health of the game character you control high in the game? Yes/No. The best answer is: \\
\textbf{Gathering own abnormal status} \\
Select the best answer to the following single-choice question based on the game-screenshot image. Respond with only the letter (Yes or No) of the correct option. Is the game character in an abnormal state? (Such as being on fire) Yes/No. The best answer is: \\
\\
\textbf{----- Comprehension -----} \\
\textbf{Understanding action intention} \\
Select the best answer to the following single-choice question based on the game-screenshot image. Respond with only the letter (Yes or No) of the correct option.Carefully observe the enemy's movements. Will the enemy attack next or is it attacking now? Yes/No. The best answer is: \\
\textbf{Understanding current state} \\
Select the best answer to the following single-choice question based on the game-screenshot image. Respond with only the letter (Yes or No) of the correct option.Is the enemy in a stunned state? (When the enemy is in a stunned state, they cannot attack for a period of time and can only be attacked. For example, the enemy is knocked down or immobilized by the spell.) Yes/No. The best answer is: \\
\\
\textbf{----- Reasoning -----} \\
Q: Select the best answer to the following single-choice question based on the game-screenshot image. Respond with only the letter (A, B, or C) of the correct option.Carefully observe the enemy's actions. As the game character, please reason which of the following actions is most suitable for your next move (ensure your health is prioritized while depleting the enemy's health). A. Restore health of the game character. B. Dodge to avoid enemy attacks and prevent damage. C. Attack the enemy. The best answer is: 
}}
\end{center}

% \begin{figure*}[h]
% \vspace{-0.8cm}
% \centering
% \includegraphics[width=1.0\textwidth]{figs/tasks.pdf}
% % \vspace{-0.8cm}
% \caption{The visualization of 13 defined tasks.}
% \label{fig:task}
% % \vspace{-0.4cm}
% \end{figure*}

\begin{figure}[!ht]
    \centering
    \includegraphics[width=1.0\linewidth]{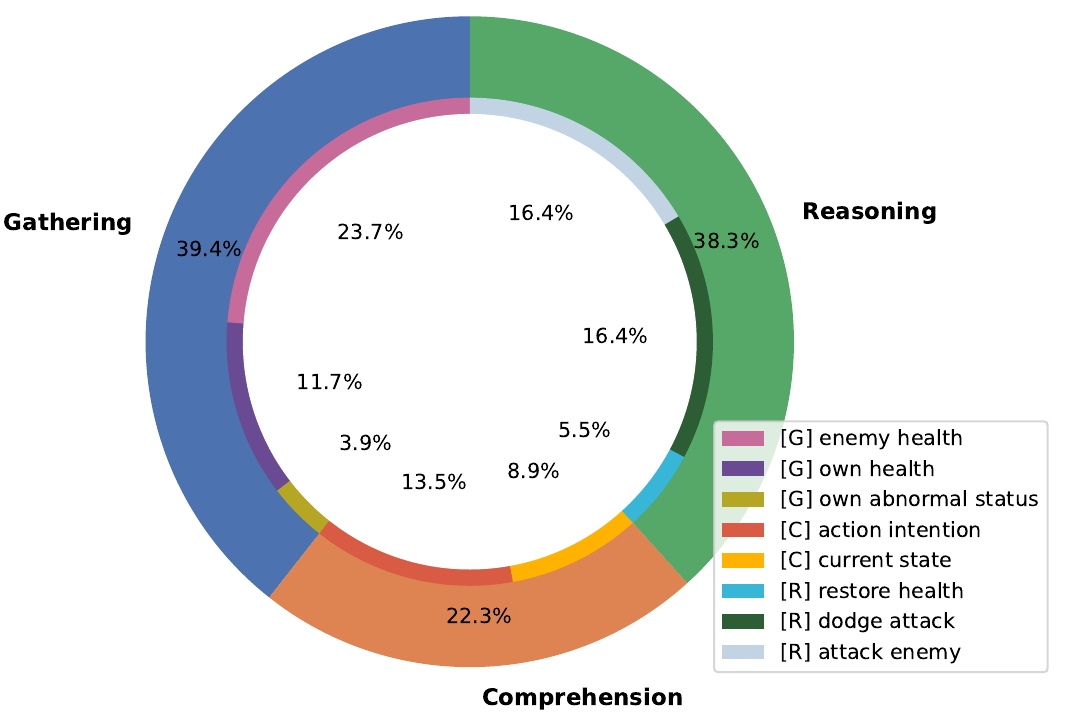}
    \vspace{-0.3cm}
    \caption{Distribution of 3 tasks (i.e., gathering, understanding, reasoning) and their 8 subtasks.}
    \label{fig:piechart}
    \vspace{-0.5cm}
\end{figure}

\subsection{Details of CUBench Benchmark} \label{sec:dataset}
To thoroughly assess the combat IQ of our CombatVLA and all baselines, we developed CUBench. As illustrated in Fig.\ref{fig:piechart}, this benchmark is composed of three types of tasks: 39.4\% gathering, 22.3\% understanding, and 38.3\% reasoning. Each of these main tasks is further divided into 8 subtasks. Tab.\ref{tab:pietable} presents a detailed breakdown.

\begin{table}[ht]
\centering
\begin{tabular}{lr}
    \toprule
    \textbf{Task Category} & \textbf{Volume} \\
    \midrule
    \textbf{Gathering} & \textbf{360} \\
    Gathering enemy health & 217 \\
    Gathering own health & 107 \\
    Gathering own abnormal status & 36 \\
    \midrule
    \textbf{Comprehension} & \textbf{204} \\
    Understanding action intention & 123 \\
    Understanding current state & 81\\
    \midrule
    \textbf{Reasoning} & \textbf{350} \\
    Option A: restore health & 50 \\
    Option B: dodge attack & 150 \\
    Option C: attack enemy & 150 \\
    \bottomrule
  \end{tabular}
\caption{Benchmark statistics of CUBench.}
\vspace{-0.4cm}
\label{tab:pietable}
\end{table}

% gathering: 360
%     收集敌人血量信息：217
%     收集自己血量信息：107
%     收集自己异常状态信息： 36
% comprehension: 204
%     理解敌人动作意图： 123
%     理解敌人目前状态： 81
% reasoning: 350
%     A选项，恢复血量：50
%     B选项，闪避躲避攻击：150
%     C选项：攻击敌人：150

\begin{figure*}[!ht]
    \centering
    \includegraphics[width=0.86\textwidth]{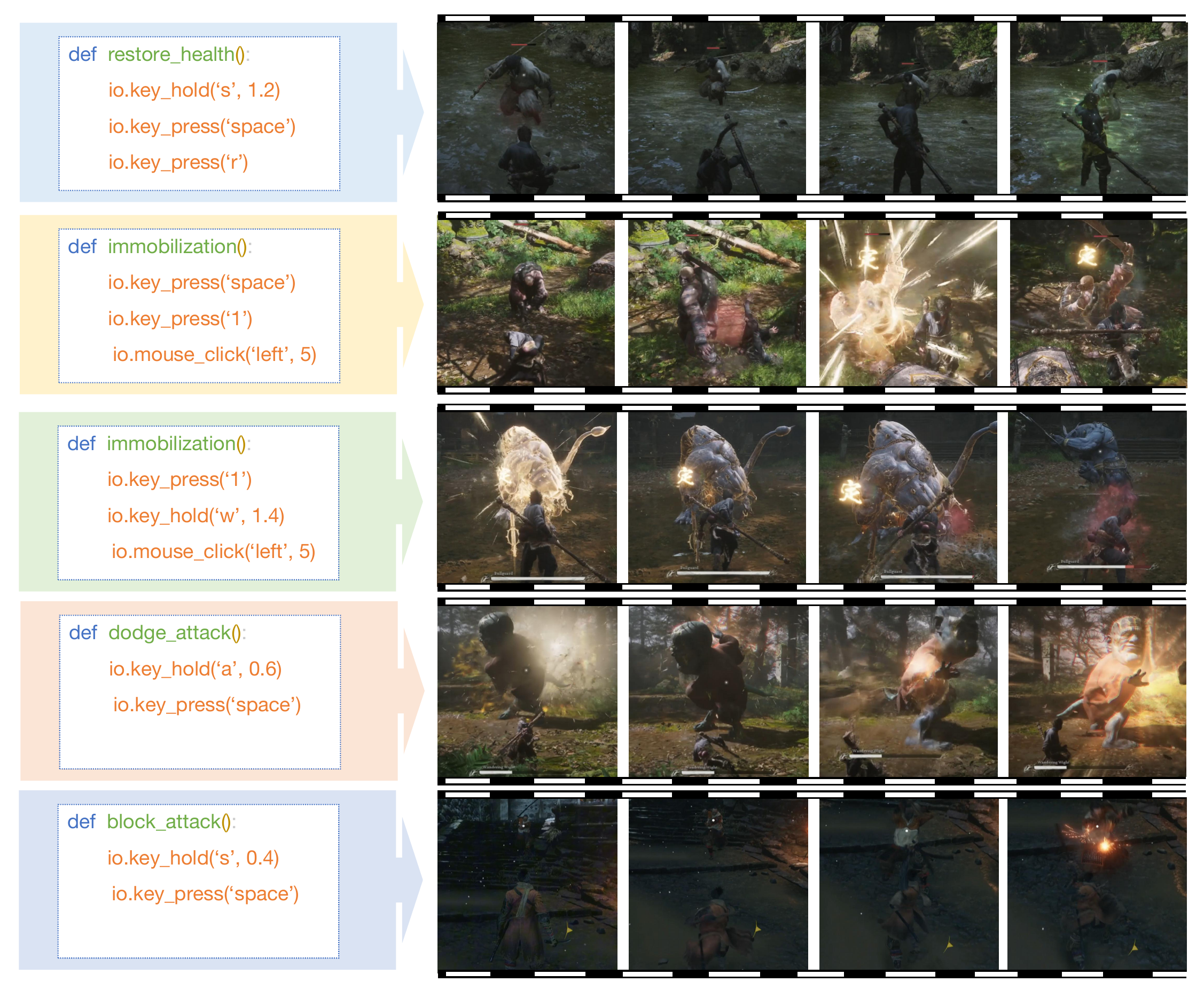}
    \vspace{-3mm}
    \caption{Additional qualitative visualization of actions and corresponding frame sequences.}
    \label{fig:add_vis}
    \vspace{-3mm}
\end{figure*}

\subsection{Details of Adaptive Action-Weighted Loss} \label{sec:loss}
In Sec.4.2 of the main content, within the `Adaptive Action-Weighted Loss' part, the predefined action sequence $P$ is [``r", ``1", ``space", ``left", ``d", ``s", ``a", ``w", ``shift", ``right"], with its corresponding weight sequence $\bm{\alpha}$ = [0.1000, 0.0549, 0.0324, 0.0211, 0.0155, 0.0126, 0.0112, 0.0105, 0.0102, 0.0100]. Since the types of actions in different ARPG games are almost the same, this predefined action sequence $P$ and $\bm{\alpha}$ have a certain level of generalizability. For instance, in the game SSDT, there is no dodge action, but there is a similar function through the block action, so you only need to change the block button in SSDT to the `space' key press.

\subsection{Details of Action Tracker} \label{sec:tracker}
The action tracker employs a multi-threaded architecture to synchronize multi-modal data collection, comprising three core technical components. First, the input monitoring module uses `pynput' to capture keyboard and mouse events with millisecond precision, distinguishing between discrete actions (such as key press/release) and continuous interactions (quantified by duration through time difference). 
Secondly, the screen capture engine utilizes `mss' for DirectX-accelerated 30 frames per second capture, combined with `OpenCV' for RGB conversion and lossless PNG compression. Using status detection mechanisms provided by `win32gui' and `psutil', recording is initiated only when the target game process (b1-Win64-Shipping.exe) occupies the foreground window, ensuring the validity of the data. Time synchronization is achieved through a unified timestamp protocol (ISO 8601 extended format with millisecond precision), and frame and action alignment is implemented during post-processing via the algorithm in Sec.3 in the main text. This alignment algorithm ensures that at the moment an action is executed, the corresponding frame does not display the game character performing the action, but rather shows it a few frames later. This allows the model to better focus on the enemy's movements.
The action tracker parses raw hardware events into basic semantic action descriptors (such as ``right mouse button held for 1.234 seconds'') and constructs the data into JSON metadata associated with the frame sequence. Experimental verification indicates event delay $\leq 15$ milliseconds, meeting the requirements for real-time interaction capture and providing high-quality AoT data for CombatVLA.

\subsection{Details of Action-of-Thought Explanation} \label{sec:aot}
After using the action tracker to collect basic data and carefully manually filtering it, we have gathered high-quality frames and actions data. Next, we define advanced semantic AoT explanations for each type of action as follows,

\begin{center}
\fcolorbox{black}{gray!10}{\parbox{1\linewidth}{
\textbf{AoT Explanation} \\
\\
\textbf{Restore Health: key `r' press} \\
The character's health is low (indicated by the white bar in the bottom left). The character needs to restore health and should create distance from enemies before healing.\\
\textbf{Immobilization: key `1' press} \\
The game character's immobilization skill is currently available. This skill can briefly freeze the enemy. It should be followed up with quick consecutive light attacks. \\
\textbf{Dodge or Block: key `space' press} \\
The enemy is about to attack the game character. The game character needs to dodge(or block in SSDT) to avoid enemy attacks and prevent damage.\\
\textbf{Light Attack: mouse `left' press} \\
The enemy is not currently attacking, so the game character should take the opportunity to execute a light attack. Consecutive uses (up to 5 times) can trigger combo moves, but they may be interrupted by enemies. \\
\textbf{Move Right: key `d' hold for n seconds} \\
The game character moves right for n seconds.\\
\textbf{Move Backward: key `s' hold for n seconds} \\
The game character moves backward for n seconds.\\
\textbf{Move Left: key `a' hold for n seconds} \\
The game character moves left for n seconds.\\
\textbf{Move Forward: key `w' hold for n seconds} \\
The game character moves forward for n seconds.\\
\textbf{Sprint: key `shift' hold} \\
The game character sprints for n seconds.\\
\textbf{Heavy Attack: mouse `right hold for n seconds} \\
The enemy is not currently attacking, so the game character can charge heavy attack for n seconds. Longer charge time increases damage but leaves vulnerable to interruption.
}}
\end{center}

\subsection{Details of Action Execution Framework} \label{sec:framework}
Our action execution framework has undergone significant modifications based on the codebase of Cradle~\cite{tan2024cradle}. Specifically, the framework records game video while the game characters execute actions, with the video recorded at a frame rate of 8 FPS and a resolution of $1920 \times 1080$. The framework samples the last 9 frames of the recorded video, evenly selecting 3 frames from these, which are resized to $1008 \times 560$ to serve as visual input for CombatVLA. Following this, CombatVLA performs inference. Due to the model's inference delay, the framework pauses the game during inference, waiting for CombatVLA to return action results before continuing the game (with an inference time of approximately 1.85 seconds). It then automatically executes the actions and records video again. Thus, we only need to call our CombatVLA \textbf{once for each inference}.

On the contrary, in terms of Cradle, performing an inference involves five processes: information gathering, self-reflection, task inference, skill curation, and action planning (with an inference time of approximately 61.68 seconds). Each of these processes requires a call to the GPT-4o model. Cradle also needs to maintain two memory libraries, which adds to the burden on the model's inference in terms of context length. Additionally, Cradle integrates the GroundingDino~\cite{liu2023grounding} model for tasks such as object detection, further increasing model inference delay and adding memory and GPU memory overhead.

\section{Additional Qualitative Visualization} \label{sec:case}
Fig.~\ref{fig:add_vis} illustrates the visualization highlights of additional combat tasks. 
In the first row, CombatVLA moves the game character away from the enemy before restoring health to ensure its own safety. 
The second and third rows show that CombatVLA charges forward to perform a series of consecutive attacks immediately after immobilizing the enemy. 
In the fourth row, the enemy's attacks can only be dodged by moving left or right or rolling, so CombatVLA first moves left and then rolls to evade. This indicates that through progressive learning, it has learned the enemy's attack patterns in task 9. 
In the fourth row, CombatVLA is able to precisely block an enemy's attack, demonstrating strong generalization capability even in zero-shot tests of different games. These cases prove that CombatVLA can make the right decisions at the right time.

\section{Task Defination} \label{sec:task}

As shown in Fig.~\ref{fig:task}, which corresponds to Tab.1 in the main text, is a visual representation of the defined tasks. The first two rows are tasks from BMW, and the last row features tasks from SSDT. The enemies in these tasks vary in appearance, attack patterns, health, and skills, which will thoroughly test the robustness of VLAs in combat tasks.

\section{Token Length of AoT} \label{sec:length}
We evaluated the average token length of AoT and truncated AoT, as shown in Tab.~\ref{tab:token}. If the model directly outputs all actions in AoT format during the inference phase, it results in an average of 73.47 redundant tokens. However, using truncated AoT can avoid this issue by only outputting the valuable action portion.

\begin{table}[]
\centering
\caption{Average token length of AoT.}
\label{tab:token}
\vspace{-0.2cm}
\begin{tabular}{cc}
\toprule
               Data Format  & Average Token Length $\downarrow$ \\
\midrule
AoT           & 116.57              \\
Truncated AoT             & 43.10               \\

\bottomrule
\end{tabular}
\vspace{-0.3cm}
\end{table}

\begin{figure*}[h]
\vspace{-0.8cm}
\centering
\includegraphics[width=1.0\textwidth]{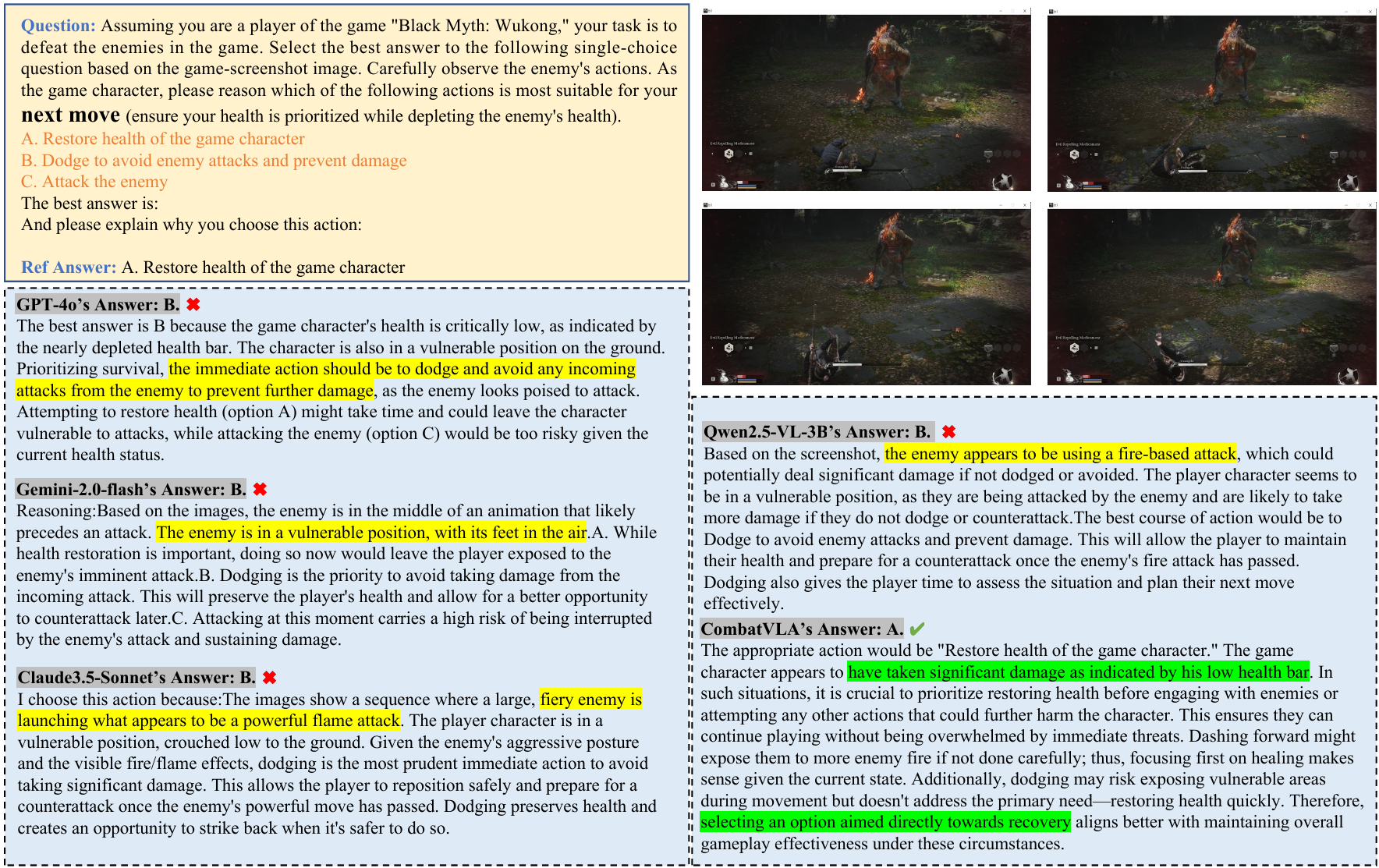}
% \vspace{-0.8cm}
\caption{Reasoning case of CUBench. The text highlighted in yellow indicates clearly incorrect reasoning, while the text highlighted in green represents correct reasoning.}
\label{fig:benchmark_vis}
% \vspace{-0.4cm}
\end{figure*}

\section{Reasoning Case of CUBench} \label{sec:cubench_case}
We demonstrated a reasoning case on CUBench, as shown in Fig.~\ref{fig:benchmark_vis}. We provided four consecutive game-screenshot frames and three action options, asking the model to infer the most appropriate next move and provide a detailed explanation.
Based on the given frames and answers, we can observe that the game character's health is critically low (the first bar in the bottom left corner of the image), while the enemy is stationary and not showing any preparatory actions for an attack. According to the hint ``prioritize self-health," the most suitable next move should be ``Option A: Restore health of the game character''.

The results show that except for CombatVLA, other models gave incorrect answers. Specifically, GPT-4o, Claude3.5-Sonnet, and Qwen2.5-VL-3B wrongly inferred that the enemy was about to attack the game character next, and therefore selected the dodge action. Gemini-2.0-flash mistakenly identified the game character as the enemy, thinking the enemy was in a vulnerable position. CombatVLA reasoning concluded that the game character had taken significant damage, indicated by his low health bar, and thus should prioritize restoring health. This case demonstrates that CombatVLA is capable of performing precise action reasoning.

\section{Demo Video} \label{sec:video}
We have provided a detailed demo video to demonstrate the effectiveness of our CombatVLA. 
The first video is a full demonstration of CombatVLA completing tasks 1 through 13. For smoother viewing, we have edited out the game pauses. The second video is a comparison of inference speeds between CombatVLA and VARP. We have kept the game pauses in this video, with the pause duration representing the inference time taken by the two methods. The video demonstrates that our method is significantly faster than VARP.
Please refer to the supplementary materials or the website \url{https://combatvla.github.io/}.

\section{All Resources.} We will open-source all resources, including the dataset, benchmark, action tracker, model weights, training code, and implementation of the framework. Due to ongoing process issues, we may gradually roll out all resources starting in April. Please allow us additional time.

% {
%     \small
%     \bibliographystyle{ieeenat_fullname}
%     \bibliography{main}
% }